\documentclass[sigconf,nonacm]{acmart}

\usepackage{multirow}
\usepackage{threeparttable}
\usepackage{graphicx}
\usepackage{textcomp}
\usepackage{xcolor}
\usepackage{enumitem}
\usepackage{comment}
\usepackage{pifont}
\usepackage{amsthm}
\usepackage{algorithm}
\usepackage{amsmath}
\usepackage{bm}
\usepackage{bbding}
\usepackage{algpseudocode}
\usepackage{fontawesome}

\newtheoremstyle{boldassumption}
  {\topsep}
  {\topsep}
  {\itshape}
  {}
  {\bfseries}
  {.}
  {.5em}
  {}

\theoremstyle{boldassumption}

\DeclareUnicodeCharacter{2217}{}

\AtBeginDocument{%
  }

\author{Chonghua~Han*}
\affiliation{
\country{Department of Electronic Engineering,  BNRist, Tsinghua University, Beijing, China}
}
\email{hanch24@mails.tsinghua.edu.cn}

\author{Yuan~Yuan*$^\dagger$}
\affiliation{
\country{Department of Electronic Engineering,  BNRist, Tsinghua University, Beijing, China}
}
\email{y-yuan20@tsinghua.org.cn}

\author{Yukun~Liu}
\affiliation{
\country{Department of Electronic Engineering,  BNRist, Tsinghua University, Beijing, China}
}
\email{liuyk21@mails.tsinghua.edu.cn}

\author{Jingtao~Ding}
\affiliation{
\country{Department of Electronic Engineering,  BNRist, Tsinghua University, Beijing, China}
}
\email{dingjt15@tsinghua.org.cn}

\author{Jie~Feng}
\affiliation{
\country{Department of Electronic Engineering,  BNRist, Tsinghua University, Beijing, China}
}
\email{fengjie@tsinghua.edu.cn}

\author{Yong~Li$^\dagger$}
\affiliation{
\country{Department of Electronic Engineering,  BNRist, Tsinghua University, Beijing, China}
}
\email{liyong07@tsinghua.edu.cn}

\renewcommand{\shortauthors}{Han et al.}

\copyrightyear{2025}
\acmYear{2025}
\setcopyright{cc}
\setcctype{by}
\acmConference[SIGSPATIAL '25]{The 33rd ACM International Conference on
Advances in Geographic Information Systems}{November 3--6,2025}{Minneapolis, MN, USA}
\acmBooktitle{The 33rd ACM International Conference on Advances in
Geographic Information Systems (SIGSPATIAL '25), November 3--6, 2025,
Minneapolis, MN, USA}
\acmDOI{10.1145/3748636.3762780}
\acmISBN{979-8-4007-2086-4/2025/11}

\begin{document}

\title{UniMove: A Unified Model for Multi-city  Human \\ Mobility Prediction}



\begin{abstract}
Human mobility prediction is vital for urban planning, transportation optimization, and personalized services. 
However, the inherent randomness, non-uniform time intervals, and complex patterns of human mobility, compounded by the heterogeneity introduced by varying city structures, infrastructure, and population densities, present significant challenges in modeling.
Existing solutions often require training separate models for each city due to distinct spatial representations and geographic coverage. 
In this paper, we propose UniMove, a unified model for multi-city human mobility prediction, addressing two challenges: (1) constructing universal spatial representations for effective token sharing across cities, and (2) modeling heterogeneous mobility patterns from varying city characteristics.  We propose a trajectory-location dual-tower architecture, with a location tower for universal spatial encoding and a trajectory tower for sequential mobility modeling. We also design MoE Transformer blocks to adaptively select experts to handle diverse movement patterns.  Extensive experiments across multiple datasets from diverse cities demonstrate that UniMove truly embodies the essence of a unified model. By enabling joint training on multi-city data with mutual data enhancement, it significantly improves mobility prediction accuracy by over 10.2\%. UniMove represents a key advancement toward realizing a true foundational model with a unified architecture for human mobility. We release the implementation at \color{blue}{\url{https://github.com/tsinghua-fib-lab/UniMove/}}.
\end{abstract}


\maketitle

\renewcommand{\thefootnote}{\fnsymbol{footnote}} 
\footnotetext[1]{Equal contribution}
\footnotetext[2]{Corresponding authors}

\section{Introduction}
\begin{figure}[t!]
    \centering
    \includegraphics[width=\linewidth]{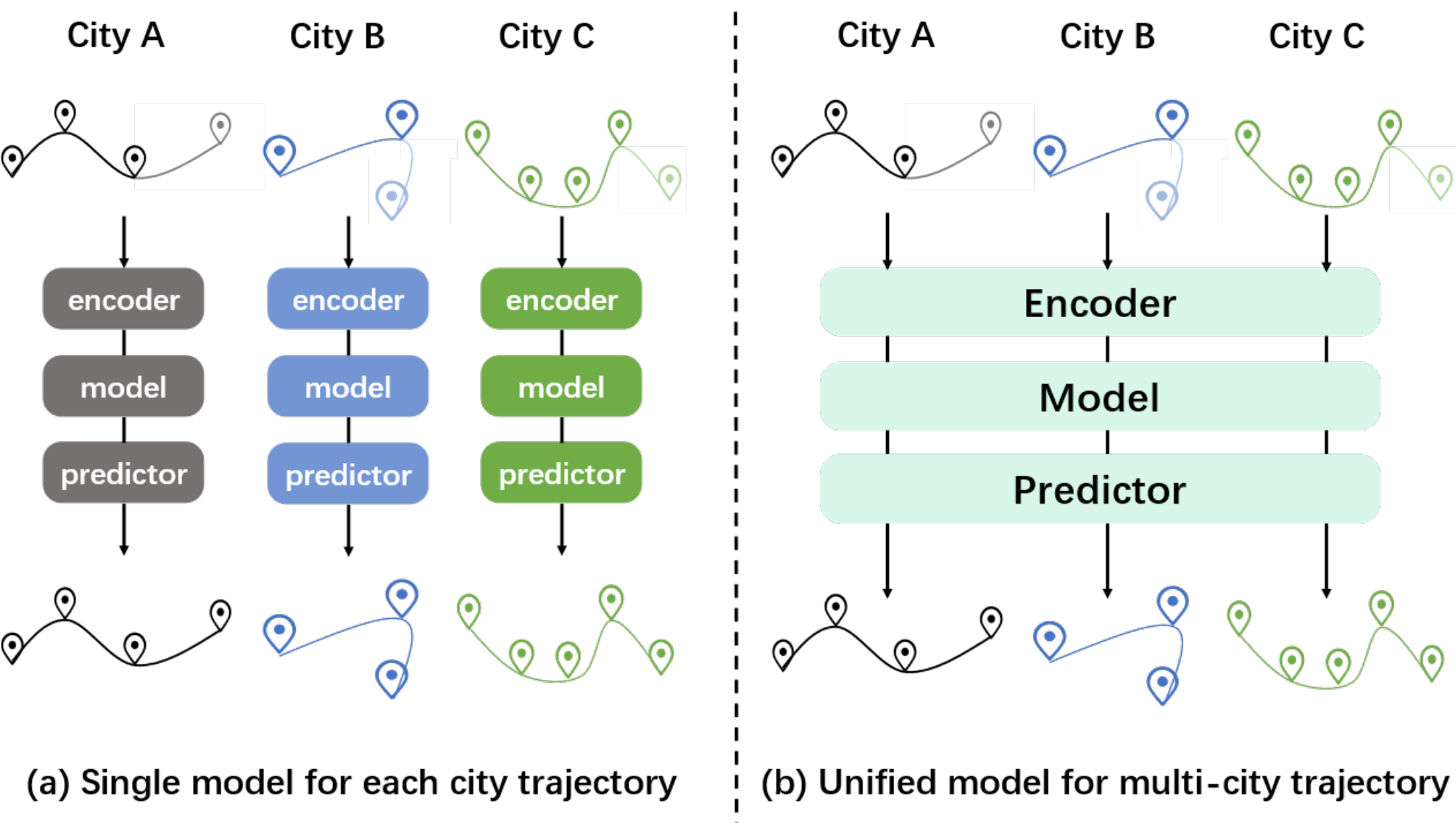}
    \caption{ Comparison between existing solutions and our approach.}
    \label{fig:trans}
\end{figure}

Human mobility~\cite{luca2021survey}, which record the movement patterns of individuals across time and space, have gained substantial importance in numerous fields, including urban planning~\cite{zheng2023spatial,sureinforcement,zheng2023road,su2024metrognn}, transportation management~\cite{zheng2024survey,yuan2024unist,yuan2024foundation}, mobile network optimization~\cite{chai2025mobiworld,sheng2025unveiling} and personalized services~\cite{lian2015content,zheng2010collaborative,yuan2023learning,zheng2011learning}. Human mobility presents unique challenges in modeling due to their inherent randomness, non-uniform time intervals, and complex patterns~\cite{yuan2025learning}. Unlike vehicle trajectories~\cite{zhu2023difftraj}, which typically follow more predictable and regular paths with time intervals on the order of seconds, human mobility data is more irregular and context-dependent.

To effectively tackle such data and harness the wealth of mobility information, researchers often treat a mobility trajectory as a sequence of location tokens, with each city represented by a set of unique tokens, similar to a vocabulary in language processing. These tokens are typically discretized using grid-based representations~\cite{long2024universal,yuan2025worldmove}, administrative divisions~\cite{yuan2025learning}, or points of interest (POIs)~\cite{feng2018deepmove,shao2024beyond}, providing a structured approach to analyzing movement patterns.  Then, advanced AI models, such as RNNs~\cite{feng2018deepmove,yuan2022activity}, transformers~\cite{si2023trajbert,long2024universal}, and other deep learning architectures~\cite{yuan2023spatio,shao2024beyond,zhu2023difftraj}, can be utilized to learn the underlying patterns and dynamics of mobility trajectories.  However, these tokens do not overlap across cities due to the unique geographic coverage of each urban area. As a result, practitioners often need to train separate models for different geographic regions, which limits generalizability and practical applicability. While some exceptions exist, such as transfer learning approaches~\cite{zhu2024controltraj,he2024st,xu2024crosspred,wang2024cola}, they either rely on vehicle trajectories (where location representations, such as numerical latitude and longitude, are transferable)~\cite{zhu2024controltraj} or just use more tokens~\cite{he2024st} or separate encoders~\cite{wang2024cola} to address geographic differences. Some methods, such as ST-MoE-BERT~\cite{stbertmoe}, force the grid cells of different cities into the same number of classes to process mobility data from different cities with a shared encoder, which severely erodes the distinct spatial identities of individual cities.

In this paper, we aim to propose a solution for training a unified model with fully shared parameters across different cities, which means that the model can accommodate the unique geographic characteristics of each urban area, while simultaneously learning shared mobility patterns that benefit from one another.  We compare our model with existing practices in Figure~\ref{fig:trans}. 
However, this is a non-trivial task, and it faces three main challenges: (1) Token sharing across cities: since each city has its own unique geographic coverage, designing a tokenizer and creating a universal spatial representation that can be shared across different cities presents a significant challenge. (2) Variations in city scale and structure: cities differ greatly in terms of scale, population density, and infrastructure. Ensuring that the model can perform well across cities with such diverse urban structures, from large metropolises to smaller towns, is a difficult task.  (3) Heterogeneity in mobility data: mobility data varies widely between cities, which presents a challenge  in designing a model that can effectively learn from these diverse data sources and enable meaningful knowledge transfer across cities.

To address the challenges above, we propose a unified mobility model \textbf{UniMove}. UniMove is a trajectory-location dual-tower architecture.  The Location Tower generates universal spatial representations based on location characteristics, utilizing these features to define locations and employing a Deep \& Cross Net to enhance their representation. The Trajectory Tower captures mobility patterns from sequential trajectories. To handle diverse movement patterns across cities, we design MoE Transformer blocks that adaptively select expert networks based on data characteristics.
Extensive experiments demonstrate that UniMove achieves superior performance compared to baseline models. Moreover, the mutual enhancement among data from multiple cities substantially boosts the model's performance, which suggests that UniMove effectively captures shared mobility patterns across cities and leverages them for mutual benefit.



In summary, our contribution can be summarized as follows:

\begin{itemize}[leftmargin=*]
\item To our best knowledge, we are the first to explore the potential for mutual enhancement among human mobility data across different cities.
\item We propose a unified human mobility prediction model \textbf{UniMove}. We propose a trajectory-location dual-tower architecture. The location tower generates universal spatial representations based on location characteristics while the trajectory tower extracts mobility patterns with MoE Transformers.
\item Extensive experiments on multiple real-world datasets demonstrate the superior performance of UniMove, which outperforms state-of-the-art baselines by 8.9\% in terms of accuracy. Meanwhile, the unified model facilitates data mutual enhancement and exhibits faster convergence.
\end{itemize}

\section{Related Work}

\subsection{Human Mobility Prediction}

Human mobility prediction has been extensively studied using various modeling techniques, each aiming to capture the intricate dynamics of human movement. Real-world applications of human mobility prediction span a variety of domains, including smart cities~\cite{wang2020review,yuan2024urbandit}, traffic management~\cite{yuan2024unist,yuan2024foundation}, personalized recommendations~\cite{lian2015content,zheng2010collaborative,zheng2011learning}, and public health (e.g., predicting the spread of diseases)~\cite{nouvellet2021reduction,han2021will,chang2021mobility}. 
Early methods relied heavily on probabilistic models, such as the Markov model~\cite{chen2014nlpmm,gambs2012next} and EPR-based models~\cite{song2010modelling,jiang2016timegeo}, where mobility patterns are derived from the transition probabilities between discrete locations. These approaches, while effective in some contexts, often fail to capture the complex, dynamic nature of human behavior~\cite{yuan2025learning,feng2018deepmove}, especially when dealing with non-uniform, irregular time intervals between events.  Recently, deep learning models have significantly advanced mobility prediction.
Models such as recurrent neural networks (RNNs)~\cite{chen2020predicting}, attention networks~\cite{yuan2023spatio,feng2018deepmove}, Transformers~\cite{corrias2023exploring, si2023trajbert}, Graph Neural Networks (GNNs)~\cite{sun2021periodicmove,terroso2022nation}, diffusion models~\cite{long2025one,zhu2023difftraj,long2024universal} have been employed to capture the temporal dependencies inherent in mobility data.
However, these models typically require training with large-scale data. 
Since mobility patterns are highly localized and influenced by unique geographic, cultural, and infrastructural factors, they necessitate training separate models for each city's data.

\subsection{Location Encoding}
Location encoding~\cite{mai2022review} plays a crucial role in human mobility prediction by transforming spatial data into a form that can be processed by deep neural networks. 
Location encoders can be divided into two main categories: continuous location encoding and discrete location encoding.  
For continuous encoding, locations are treated as continuous values, often represented by continuous coordinates (e.g., latitude, longitude)~\cite{zhu2024unitraj,zhang2024noise,yuan2022activity,yuan2024generating}. These continuous values can be directly embedded into the model through continuous embeddings~\cite{cao2021generating,zhu2024unitraj,zhang2024noise} or positional encodings~\cite{yuan2022activity,yuan2023spatio}, which are typically processed via neural networks to capture spatial relationships in a continuous space.
In discrete location encoding, the continuous spatial domain is often divided into predefined locations, such as Points of Interest (POI)~\cite{feng2018deepmove,shao2024beyond}, Areas of Interest (AOI)~\cite{zhang2023city}, administrative districtc~\cite{yuan2025learning}, or grids~\cite{long2024universal,yuan2025learning}. Each of these locations is assigned a discrete identifier, and the locations are then represented using techniques like one-hot encoding~\cite{brownlee2020ordinal}. In this way, each location is transformed into a unique token, much like words in language models. Discrete encoding simplifies the representation, making it more computationally efficient while still allowing the model to process and distinguish between different locations effectively.

\subsection{Cross-City Modeling of Human Mobility}
Cross-city modeling of human mobility aims to leverage knowledge and patterns learned from one city to improve mobility modeling in another, addressing the challenges posed by data scarcity and the unique characteristics of individual cities~\cite{he2020human,xu2024crosspred,wang2024cola,jiang2021transfer,gupta2022doing}. 
For instance, He \textit{et al.}~\cite{he2020human} propose a top-down paradigm that utilizes abundant annotated data to plan mobility for a new city. Additionally, meta-learning frameworks~\cite{vilalta2002perspective} have been developed to enable models to adapt quickly to new cities with limited data~\cite{gupta2022doing,wang2024cola}. However, existing approaches still face significant limitations. For example, they often fail to address the challenge of using a single shared model (with identical parameters) across different cities. Current methods either rely on separate location embedding layers for each city~\cite{wang2024cola} or are restricted to handling vehicle trajectories with numerical coordinate values~\cite{zhu2024controltraj,zhu2024unitraj}.
Unlike vehicle trajectories, which primarily rely on numerical GPS coordinates (e.g., latitude and longitude), human mobility trajectories are inherently richer in semantics, encompassing contextual information such as PoIs, activities, and urban infrastructure. This requires to address the tokenization of locations across cities, ensuring that models can effectively generalize and adapt to diverse urban environments. 


\section{Preliminary}

\begin{figure*}[t!]
    \vspace{-3mm}
    \centering
    \includegraphics[width=\linewidth]{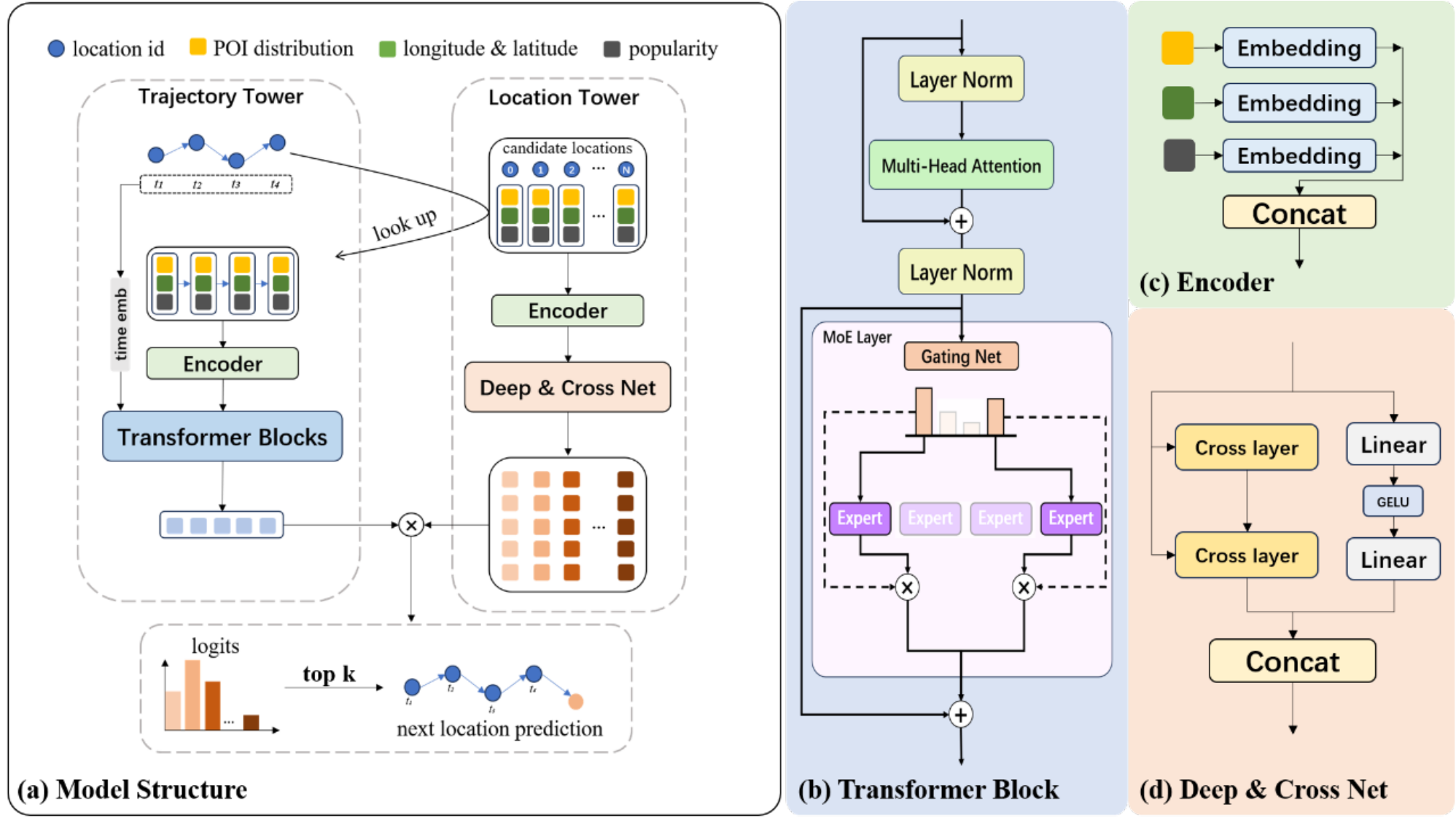}
    \caption{The overall architecture of UniMove. (a)Trajectory Tower and Location Tower:Trajectory Tower predicts the latent embedding of the next location based on historical trajectories, Location Tower generates the feature representations for all candidate locations. (b)MoE(Mixture of Experts) Transformer Block. (c)Location Encoder. (d)Deep \& Cross Net for further capturing features from different locations.}
    \label{fig:model}
    \vspace{-3mm}
\end{figure*}

\subsection{Human Mobility Data}
Human mobility datasets are non-uniform trajectories with variable lengths. The relevant definitions are provided as follows:

\textbf{Definition of Location:} A location is a grid area within a city, uniquely identified by a specific location ID. The characteristics of each location include its geographical coordinates, the distribution of Points of Interest (POI), and popularity. The latitude and longitude coordinates pinpoint the exact location of the area within the city. The distribution of POIs reflects the functional nature of the area. The popularity, measured by the number of visits during a certain period, is a key indicator of its appeal and activity level.

\textbf{Definition of Mobility Trajectory:} The mobility trajectory of a user, denoted as \( S_u \), can be represented as \(\{ \text{loc}_1, t_1; \text{loc}_2, t_2; \ldots; \text{loc}_n, t_n \}\), where \(\text{loc}_i\) represents the location ID where the user stays, and \( t_i \) represents the start time when the user first appears at location \(\text{loc}_i\). Since precise time information at the millisecond level is redundant and even noisy for our task, we divide a day into 48 time slots at half-hour intervals and map \( t_i \) to the nearest discrete time point. As each user's mobility pattern is different, although the trajectory sequences are within the same time window \( T \), the length \( n \) of each sequence is not fixed, and the time interval between adjacent locations \( t_i \) and \( t_{i+1} \) in each trajectory is also not fixed.

\subsection{Mobility Prediction}
In the context of mobility prediction, a typical scenario involves leveraging a variable number of historical records to forecast the subsequent future location. Specifically, given a sequence of historical mobility trajectories denoted as $\{S_t\}_{t=1}^H$, where $H$ represents the length of the historical time steps and is not fixed, the objective is to predict the future location $loc_{H+1}$, This task can be formally defined as learning a mapping function $\mathcal{F}$ that transforms the historical sequence into the future location, which can be mathematically represented as:
\begin{equation}
\hat{loc}_{H+1} = \mathcal{F}(S_{1:H})
\end{equation}
where $\hat{loc}_{H+1}$ signifies the predicted location for the next stay point, and $S_{1:H}$ denotes the historical records spanning the time period from 1 to $H$. The primary goal is to accurately predict the next future location based on the historical data provided.

The mobility prediction problem bears resemblance to the next token prediction in the close task of natural language processing(NLP). However, unlike NLP, where multiple datasets can utilize the same vocabulary, each city's mobility trajectory dataset varies in the number and features of locations. When applied to multiple datasets, existing methods require different encoders and predictors for each city, and train the model from scratch on the new dataset. It can be represented as:
\begin{equation}
\hat{loc}_{H+1} = \mathcal{F}_i(S_{1:H}) \quad  S \in \mathcal{D}_i  , \mathcal{D} = \{ \mathcal{D}_1 , \mathcal{D}_2, \ldots \}
\end{equation}
where $\mathcal{D}$ is a collection of mobility trajectory datasets from multiple cities.
Our motivation is to develop a prediction model that can be applied to datasets from multiple cities without any modification, which can be mathematically represented as:
\begin{equation}
\hat{loc}_{H+1} = \mathcal{F}(S_{1:H}) \quad  S \in \mathcal{D} , \mathcal{D} = \{ \mathcal{D}_1, \mathcal{D}_2, \ldots \}
\end{equation}
\section{Methodology}

\begin{table}[t!]
\centering

\caption{Details of POI Categories}
\resizebox{0.7\columnwidth}{!}{
\begin{tabular}{|l|}
\hline
Auto Service  \\
\hline
Auto Dealers   \\
\hline
 Auto Repair\\
\hline
Motocycle Service\\
\hline
Food \& Beverages  \\
\hline
Shopping  \\
\hline
 Daily Life Service \\
\hline
Sports \& Recreation\\
\hline
Medical Service\\
\hline
Accommodation Service \\
\hline
 Tourist Attraction\\
\hline
Commercial House\\
\hline
Governmental Organization\\
 \hline
 Science/Culture \& Education Service  \\
 \hline
\end{tabular}
}
\label{poi}

\end{table}

\begin{table}[t!]
\centering

\caption{Details of Popularity Rank and Value}
\resizebox{0.6\columnwidth}{!}{
\begin{tabular}{|c|c|}
\hline
\textbf{Popularity Rank} & \textbf{R value} \\ \hline
<1\% & 0 \\ \hline
1\%-5\% & 1 \\ \hline
5\%-10\% & 2 \\ \hline
10\%-20\% & 3 \\ \hline
20\%-40\% & 4 \\ \hline
40\%-60\% & 5 \\ \hline
60\%-80\% & 6 \\ \hline
>80\% & 7 \\ \hline
\end{tabular}
}
\label{popularity rank}
\end{table}

\subsection{Overall Framework}
We propose a unified mobility prediction framework UniMove, which adopts a dual-tower structure, as shown in the Figure~\ref{fig:model}. The Trajectory Tower predicts the intent embedding $\mathbf{I} \in \mathbb{R}^{B \times T \times d}$ of the next location based on the user's historical trajectories. The Location Tower generates candidate locations embedding $\mathbf{L} \in \mathbb{R}^{N \times d}$ based on the characteristics of locations within the city. the predicted location probability distribution $\mathbf{M} \in \mathbb{R}^{B \times T \times N}$ is obtained by element-wise multiplying $\mathbf{L}$ and $\mathbf{I}$. The framework comprises three main components:
\begin{itemize}[leftmargin=*]
\item \textbf{Location Encoder}:The Location Encoder takes the regional features of a location as input. Compared to existing methods that directly use location IDs as feature inputs, this approach can be applied to extract features from locations in different cities without altering any parameters
\item \textbf{Deep \& Cross Net}:We introduce DCN in the Location Tower to model the features of locations within the city. This approach not only enables automatic learning of complex interactions between features but also further extracts unique characteristics of different locations through feature crossing, thereby providing more complex feature representations for subsequent prediction tasks.
\item  \textbf{MoE Transformer Block}: To address the heterogeneous distribution of mobility data between different cities, UniMove utilize MoE transformer, which assigns different data distributions to specialized expert models. This reduces conflicts caused by data heterogeneity and improves the model's generalization capability.
\end{itemize}

\subsection{Location Encoder}

Each location is defined as a grid area, and the location embedding is obtained from three features: POI distribution $\mathbf{P}$, geographical coordinates $\mathbf{G}$, and popularity rank $\mathbf{R}$. 
\begin{itemize}[leftmargin=*]
\item POI distribution$\mathbf{P}\in \mathbb{R}^{2c}$, where c represents the number of POI categories. The details of POI categories are shown in Table \ref{poi}. The POI distribution is encoded into the form $[n_1, n_2, \dots, n_{c}, p_1, p_2, \dots, p_{c}]$, where $n_i$ denotes the number of POIs in the $i$-th category, and  \( p_i = \frac{n_i}{\sum_{j=1}^{c} n_j} \)
\item Geographical coordinates$\mathbf{G}\in \mathbb{R}^{2}$ refer to the latitude and longitude of the location's center point.
\item $\mathbf{R}\in \mathbb{R}^{1}$ is a discrete integer value.Based on the flow data of all locations within the city, each location is ranked according to its popularity.The $\mathbf{R}$ value is assigned based on the percentile ranking: locations in the top $1\%$ are assigned a value of 1, those between $1\%$ and $5\%$ are assigned a value of 2, and so on.The details of popularity rank and value are shown in Table \ref{popularity rank}.
\end{itemize}

We then create three embedding layers to obtain the following embeddings: POI embedding $\mathbf{E}_p \in \mathbb{R}^{B \times T \times \frac{d}{2}}$, geographical embedding $\mathbf{E}_g \in \mathbb{R}^{B \times T \times \frac{d}{4}}$, and flow rank embedding $\mathbf{E}_r \in \mathbb{R}^{B \times T \times \frac{d}{4}}$. The location embedding $\mathbf{E}_l \in \mathbb{R}^{B \times T \times d}$ is defined as the concatenation of these three embeddings: 
\begin{equation}
\mathbf{E}_l = \text{concat}(\mathbf{E}_p, \mathbf{E}_g, \mathbf{E}_r)
\end{equation}
\subsection{Deep \& Cross Net }
Location Tower introduces the Deep \& Cross Net(DCN) to further capture features from different locations. It consists of two main parts: the Cross layer and the Deep layer.
The design of the Cross layer is inspired by ~\cite{wang2017deep}, aiming to capture high-order feature interactions through feature crossing. 
Suppose the input feature vector is location embedding $\mathbf{E_l}$, with a dimension of $d$. The Cross layer calculates as follows:
\begin{equation}
\mathbf{E}_{\text{cross}} = \sum_{i=1}^{d} \mathbf{E}_i \odot \mathbf{W}_i \mathbf{E}_l + \mathbf{b}_i
\end{equation}
where $\odot$ represents element-wise multiplication, and $\mathbf{W}_i$ is the learnable weight matrix. In this way, the Cross layer can effectively capture the nonlinear relationships between features.
The Deep layer further processes features through a multi-layer perceptron (MLP) structure. Its calculation method is as follows:
\begin{equation}
\mathbf{E}_{\text{deep}} = (\text{GELU}(\mathbf{W}_1 \mathbf{E_l} + \mathbf{b}_1)) \mathbf{W}_2 + \mathbf{b}_2
\end{equation}
where $\mathbf{W}_1$ and $\mathbf{W}_2$ are learnable weight matrices, $\mathbf{b}_1$ and $\mathbf{b}_2$ are bias terms, and GELU is an activation function that introduces nonlinearity.
The candidate locations embedding $\mathbf{L} \in \mathbb{R}^{N \times d}$ is obtained by concatenating the outputs of the Cross layer and the Deep layer. The specific expression is as follows:
\begin{equation}
\mathbf{L} = \text{Concat}(\mathbf{E}_{\text{cross}}, \mathbf{E}_{\text{deep}})
\end{equation}
\subsection{MoE Transformer Blocks}
The Trajectory Tower is composed of MoE transformer blocks that integrate multi-head masked attention and Mixture-of-Experts (MoE) layers. It is designed to model mobility patterns and predict the intent embedding $\mathbf{I}$ for the next location based on the user's historical trajectory. This process can be formulated as:
\begin{equation}
    \mathbf{I} = \text{Transformer}(\mathbf{E}_l + \mathbf{E}_t),
\end{equation}
where $\mathbf{E}_l$ represents the embedding of the user's historical trajectory and $\mathbf{E}_t\in \mathbb{R}^{B \times T \times d}$ denotes the embedding derived from the timestamp information at each trajectory point.
\subsubsection{Masked multi-head attention}
The masked multi-head attention mechanism in our model incorporates two types of masks to ensure proper information flow and handle variable-length trajectories:
\begin{itemize}[leftmargin=*]
\item \textbf{Causal Mask}: The causal mask ensures that the prediction at the \(i_{\text{th}}\) location is constrained to utilize information solely from the first \(i-1\) locations. Specifically, it enforces a unidirectional dependency structure, where each location is influenced exclusively by its preceding locations. This design is essential for modeling sequential patterns in trajectories, as it effectively prevents the leakage of future information and preserves the temporal order inherent in the trajectory data.
\item \textbf{Padding Mask}: Since the lengths of user trajectories vary, we standardize the input length for the model by padding all trajectories to a fixed length of \(T\). To achieve this, we append a special termination token (represented by 1) at the end of each trajectory to signify its termination point. Subsequently, the remaining positions are filled with padding tokens (represented by 0) to reach the desired length. This padding strategy ensures that the model can process trajectories of different lengths while maintaining the integrity of the original data structure.
\end{itemize}
\subsubsection{Mixture of Experts (MoE)}
The MoE is composed of a gating network and multiple expert networks, which serve as an alternative to the Feed-Forward Network (FFN) in the Transformer architecture.
The gating network computes the activation probabilities for each expert based on the input $x$, producing a probability distribution vector $G(x)$. The output of the gating network is calculated as follows:
\begin{equation}
G(x) = \text{Softmax}(\text{TopK}(x \cdot W_g))
\end{equation}
where $W_g$ is the weight matrix of the gating network, and $\text{TopK}(\cdot)$ denotes the operation of selecting the top $k$ largest values from the input vector, with the remaining values set to $-\infty$ to ensure that only the top $k$ experts are activated.

During the training of MoE model, it is common for the model to favor a single expert, leading to load imbalance among the experts. To address this issue, noise is added to the output of the gating neural network. This helps to balance the load among different experts by preventing the model from consistently routing inputs to the same expert. Specifically, standard normal noise is added to the logits produced by the gating network.This noise is scaled by a factor derived from a separate linear layer, which allows for adjustable noise levels, which can be expressed as follows:
\begin{equation}
\text{noise} = \mathcal{N}(0, 1) \times \text{Softplus}(x \cdot W_n)
\end{equation}
where \(\mathcal{N}(0, 1)\) represents the standard normal noise, and ${W}_n$ is the weight matrix of the additional noise linear layer. The noisy logits are then used to select the top-k experts, ensuring that the selection process is more randomized and less biased towards any particular expert. So the final logits of the gating network can be expressed as
\begin{equation}
G(x) = \text{Softmax}(\text{TopK}(x \cdot W_g + \text{noise}))
\end{equation}

This method not only promotes load balancing but also enhances the model's robustness and generalization ability.
Each expert network produces an output $E_i(x)$, where $i$ indicates the $i$-th expert. For each input $x$, only the $k$ selected experts by the gating network will compute their outputs.

Ultimately, the output of the MoE layer $H_{\text{moe}}$ is the weighted sum of the outputs from the selected $k$ experts, with the weights determined by the gating network's output:
\begin{equation}
H_{\text{moe}} = \sum_{i=1}^{k} G_i(x) \cdot E_i(x)
\end{equation}

This sparse activation mechanism ensures that only a subset of experts participate in the computation, the dynamic selection by the gating network allows the model to choose the most suitable experts based on the characteristics of the input, enhancing the model's adaptability and generalization capability.
\subsection{Training}

\begin{algorithm}[t]
\caption{Training}\label{alg:train}

\begin{algorithmic}[1]
\Require Trajectory Dataset $D = \{D_1, D_2, \ldots, D_M\}$, Mobility prediction model $F$, and loss function $L$.
\Ensure Learnable parameters $\theta$ for the model $F$
\For{$epoch\in\{1,2,\ldots,N_{iter}\}$}
\State Randomly sample a mini-batch $x$ from $D_m$. 
\State Trajectory data $x_i = (P_i, G_i, R_i, t_i)$.
\State location data $l = (P^N, G^N, R^N)$,
\State $E_p, E_g, E_r, E_t \gets \text{emb}(P_i), \text{emb}(G_i), \text{emb}(R_i), \text{emb}(t_i)$ 
\State $I = \text{Transformer}(\text{concat}(E_p, E_g, E_r)+E_t)$
\State $M_p, M_g, M_r \gets \text{emb}(P^N), \text{emb}(G^N), \text{emb}(R^N)$ 
\State $L = \text{DCN}(\text{concat}(M_p, M_g, M_r))$
\State logits = $I \odot L$
\State Compute the Cross Entropy loss $\mathcal{L} \leftarrow L(logits,x)$
\State Update the model's parameters $\theta \leftarrow update(\mathcal{L};\theta)$
\EndFor
\end{algorithmic}
\end{algorithm}
As summarized in the algorithm ~\ref{alg:train}, during the training process, a batch of trajectory data is extracted from the dataset of a random city. The user's historical trajectory data is fed into the Transformer-based Trajectory Tower to derive the intent embedding$\mathbf{I} \in \mathbb{R}^{B \times T\times d}$for the next location prediction. Concurrently, the features of all locations within the city are input into the Location Tower to generate the location embedding $\mathbf{L} \in \mathbb{R}^{N \times d}$. The final $\mathbf{logits} \in \mathbb{R}^{B \times T\times N}$ is computed through element-wise multiplication of L and I, with the logits calculation being as follows:
\begin{equation}
\text{logits}_{b,t,n} = \sum_{i=1}^{d} I_{b,t,i} \cdot L_{n,i} 
\end{equation}
and the loss is expressed as:
\begin{equation}
Loss = \text{Cross-Entropy Loss}(\text{logits},x)
\end{equation}
where $\mathbf{x} \in \mathbb{R}^{B \times T}$ is the trajectory location id.

\section{Performance Evaluations}

\subsection{Experimental Settings}

\subsubsection{Datasets}
We conduct extensive experiments on three real-world mobility datasets: Shanghai, Nanchang, and Lhasa. Detailed information on the data sets is summarized in Table~\ref{tbl:dataset_info}. For pre-processing the trajectory data in these datasets, we applied a sliding window of three days to each user's trajectory and filtered out trajectories with fewer than five trajectory points. In the location pre-processing phase, we map GPS points to predefined grid area IDs with a granularity of $500\,\text{m} \times 500\,\text{m}$ and record features such as the POI distribution within each grid area. For every location, POI data is sourced from Gaode Map and consists of 14 major categories. The longitude and latitude of all locations in the same city are normalized to have a mean of 0 and a standard deviation of 1. We discretize the popularity rank based on quantiles. For temporal pre-processing, we organized the time data into fixed half-hourly intervals. Finally, we divided the data into training, validation, and testing sets based on different users, with a ratio of $6:2:2$.
\begin{table}[H]
\caption{Basic statistics of mobility data.}
\label{tbl:dataset_info}
\begin{threeparttable}
\resizebox{0.8\columnwidth}{!}{
\begin{tabular}{cccc}
\toprule
City &  Duration & Location & Trajectory\\
\hline
Shanghai & 14 days & 5451 & 483200 \\
Nanchang & 7 days & 2055 & 26400 \\
Lhasa & 28 days & 166 & 48160 \\
\bottomrule
\end{tabular}}
\end{threeparttable}
\end{table}

\begin{table*}[t!]
\caption{Performance comparison of different prediction models in terms of Acc@k, where bold denotes best results. }
\label{tbl:result}
\begin{threeparttable}
\resizebox{2\columnwidth}{!}{
\begin{tabular}{cccccccccc}
\toprule
& \multicolumn{3}{c}{\textbf{Nanchang}} & \multicolumn{3}{c}{\textbf{Lhasa}} & \multicolumn{3}{c}{\textbf{Shanghai}} \\
\cmidrule(lr){2-4} \cmidrule(lr){5-7} \cmidrule(lr){8-10}
\textbf{Model} & \textbf{Acc@1} & \textbf{Acc@3} & \textbf{Acc@5}& \textbf{Acc@1} & \textbf{Acc@3} & \textbf{Acc@5}& \textbf{Acc@1} & \textbf{Acc@3} & \textbf{Acc@5} \\
\cmidrule(lr){1-1} \cmidrule(lr){2-4} \cmidrule(lr){5-7} \cmidrule(lr){8-10}
Linear &0.125	&0.300 &0.398&0.253&0.500&0.622&0.105&0.239&0.326 \\
Markov &0.166&0.353 &0.402&0.288&0.557&0.688&0.154&0.303&0.406  \\
LSTM &0.224&0.416&0.541&0.366 &0.601 &0.702&0.206&0.353&0.459 \\
Transformer &0.225&0.448&0.543&0.380&0.614&0.715&0.232&0.404&0.454 \\
DeepMove &0.243&0.420&0.503&0.401&0.610&0.690&0.214&0.365&0.445 \\
TrajBert &0.249&0.423&0.505&0.353&0.571&0.672&0.246&0.411&\textbf{0.489} \\
CTLE &0.252 &0.435 &0.508 & 0.413 & 0.588& 0.682&0.260&0.405&0.445\\
STAN &0.244 & 0.422 &0.507 & 0.384 & 0.600& 0.685&0.247&0.409&0.469\\
GETNext &0.258 &0.438 &0.522& 0.408 & 0.619& 0.713&0.262&0.400&0.477\\
\cmidrule(lr){1-1} \cmidrule(lr){2-4} \cmidrule(lr){5-7} \cmidrule(lr){8-10}
\textbf{UniMove}(separate)&0.261&0.455&0.506&0.421&0.557&0.679&0.244&0.405&0.462\\
\textbf{UniMove} &\textbf{0.338}	&\textbf{0.509}	&\textbf{0.573} &\textbf{0.425} & \textbf{0.646}&\textbf{0.722}&\textbf{0.263}&\textbf{0.412}&0.476\\

\bottomrule
\end{tabular}}
\end{threeparttable}
\end{table*}
\begin{table*}[t]
\caption{Results of the ablation study.}
\label{tbl:ablation}
\begin{threeparttable}
\resizebox{1.8\columnwidth}{!}{
\begin{tabular}{cccccccccc}
\toprule
& \multicolumn{3}{c}{\textbf{Nanchang}} & \multicolumn{3}{c}{\textbf{Lhasa}} & \multicolumn{3}{c}{\textbf{Shanghai}} \\
\cmidrule(lr){2-4} \cmidrule(lr){5-7} \cmidrule(lr){8-10}
\textbf{Model} & \textbf{Acc@1} & \textbf{Acc@3} & \textbf{Acc@5}& \textbf{Acc@1} & \textbf{Acc@3} & \textbf{Acc@5}& \textbf{Acc@1} & \textbf{Acc@3} & \textbf{Acc@5} \\
\cmidrule(lr){1-1} \cmidrule(lr){2-4} \cmidrule(lr){5-7} \cmidrule(lr){8-10}
w/o dcn & 0.316 & 0.492 & 0.560 & 0.414 & 0.632 & 0.720 & 0.243 & 0.379 & 0.421 \\
w/o moe & 0.321 & 0.497 & 0.562 & 0.405 & 0.617 & 0.705 & 0.261 & 0.403 & 0.461 \\
\textbf{UniMove} & \textbf{0.338} & \textbf{0.509} & \textbf{0.573} & \textbf{0.425}& \textbf{0.646}& \textbf{0.722} & \textbf{0.263} & \textbf{0.412} & \textbf{0.476} \\
\bottomrule
\end{tabular}}
\end{threeparttable}
\end{table*}

\subsubsection{Metrics}
Evaluation metric Acc@K is utilized to measure the proportion of samples that are correctly predicted within the top \( K \) highest-probability locations by the model.The formula for Acc@k can be expressed as follows:
\begin{equation}
    \text{Acc}@k = \frac{1}{N} \sum_{i=1}^N \mathbb{I}\left( y_i \in \{ f(x_i)_1, f(x_i)_2, \ldots, f(x_i)_k \} \right)
\end{equation}
where $\mathbb{I}(\cdot)$ is the indicator function (1 if the condition is true, 0 otherwise), $N$ is the total number of samples, $y_i$ is the true label for the $i$-th sample, $f(x_i)_1, \ldots, f(x_i)_k$ are the top $k$ predictions made by the model for the sample $x_i$.

\subsubsection{Baselines}
To evaluate the performance of our method, we compared it with state-of-the-art models.The details are as follows:
\begin{itemize}[leftmargin=*]
\item \textbf{Linear}:Linear model employs the linear regression to predict the probability distribution of the next location.
\item \textbf{Markov}~\cite{gambs2012next}:The Markov model serves as a statistical framework designed to depict how states evolve over time. It predicts future locations by computing the probabilities of transitions.
\item \textbf{LSTM}~\cite{kong2018hst}:LSTM networks excel at processing sequential data and are adept at capturing long-term dependencies, making them naturally suitable for location prediction tasks.
\item \textbf{Transformer}~\cite{vaswani2017attention}:Transformer is a deep learning architecture based on the self-attention mechanism, capable of efficiently processing sequential data and capturing long-range dependencies.
\item \textbf{DeepMove}~\cite{feng2018deepmove}:This approach is particularly effective in addressing the challenges of predicting human movement patterns, which are influenced by various factors such as time, location, and personal preferences
\item \textbf{TrajBert}~\cite{si2023trajbert}:TrajBERT is a trajectory recovery method based on BERT, which recovers implicit sparse trajectories through spatial-temporal refinement.
\item \textbf{STAN}~\cite{luo2021stan}: STAN employs a spatio-temporal bi-attention architecture to capture non-adjacent location correlations and non-consecutive visit dependencies. 
\item \textbf{GETNext}~\cite{yang2022getnext}: GETNext integrates Graph Convolutional Networks (GCN) and Transformer to model global POI transition patterns via a trajectory flow map. It fuses spatio-temporal contexts with time-aware category embeddings for next POI recommendation.
\item \textbf{CTLE}~\cite{lin2021pre}: The Context and Time aware Location Embedding (CTLE) model dynamically generates context-specific location embeddings using bidirectional Transformers and temporal encoding to capture multi-functional properties of locations under varying contexts and temporal dynamics.

\end{itemize}

\subsubsection{Implementation Details}
The transformer blocks utilized in this study comprises 8 layers, with an embedding dimension of 512. The MoE layer within the model consists of 4 experts and employs a top-$k$ gating mechanism, where $k=2$. Training was conducted using the AdamW optimizer, with a learning rate set to $3 \times 10^{-4}$, for a total of 50 epochs. Early stopping was implemented to terminate training if the validation loss did not improve for 3 consecutive epochs. The batch size was fixed at 16.

\subsection{Overall Performance}

The overall performance of UniMove is compared with that of the baselines, and the results are presented in Table~\ref{tbl:result}. Based on the experimental findings, we have reached the following conclusions:
\begin{itemize}[leftmargin=*]
\item \textbf{UniMove outperforms the state-of-the-art baselines in the mobility prediction task.} By integrating the trajectory data of multiple cities for joint training, UniMove can more accurately capture the complex patterns of human mobility, thereby effectively improving the model's prediction performance. Compared with the SOTA baselines, the accuracy of UniMove has been significantly improved, with the highest increase reaching 35.7\% in terms of Acc@1 for Nanchang dataset.
\item \textbf{Data from different cities can complement each other, thereby enhancing the performance of UniMove.} Compared with training UniMove solely on a single dataset, incorporating mobility trajectory data from other cities can significantly enhance the model's prediction accuracy for target city trajectories. Specifically, across multi-city datasets, the performance metrics ACC@1, ACC@3, and ACC@5 have improved by 12.7\%, 9.8\%, and 7.5\%, respectively.
\item \textbf{The performance improvement of UniMove is more pronounced in the city with sparse mobility data.} For Nanchang city with limited data and incomplete mobility patterns, UniMove can effectively leverage the data resources from data-rich cities with complete mobility patterns, such as Shanghai.UniMove achieves better performance improvements on the sparse Nanchang dataset than on the Shanghai dataset. Specifically, on the Shanghai dataset, UniMove achieves an average improvement of 18.1\%, while on the Nanchang dataset, the metrics improved by an average of 4.1\%.
\end{itemize}

\begin{figure*}[t!]
    \centering
    \includegraphics[width=1\linewidth]{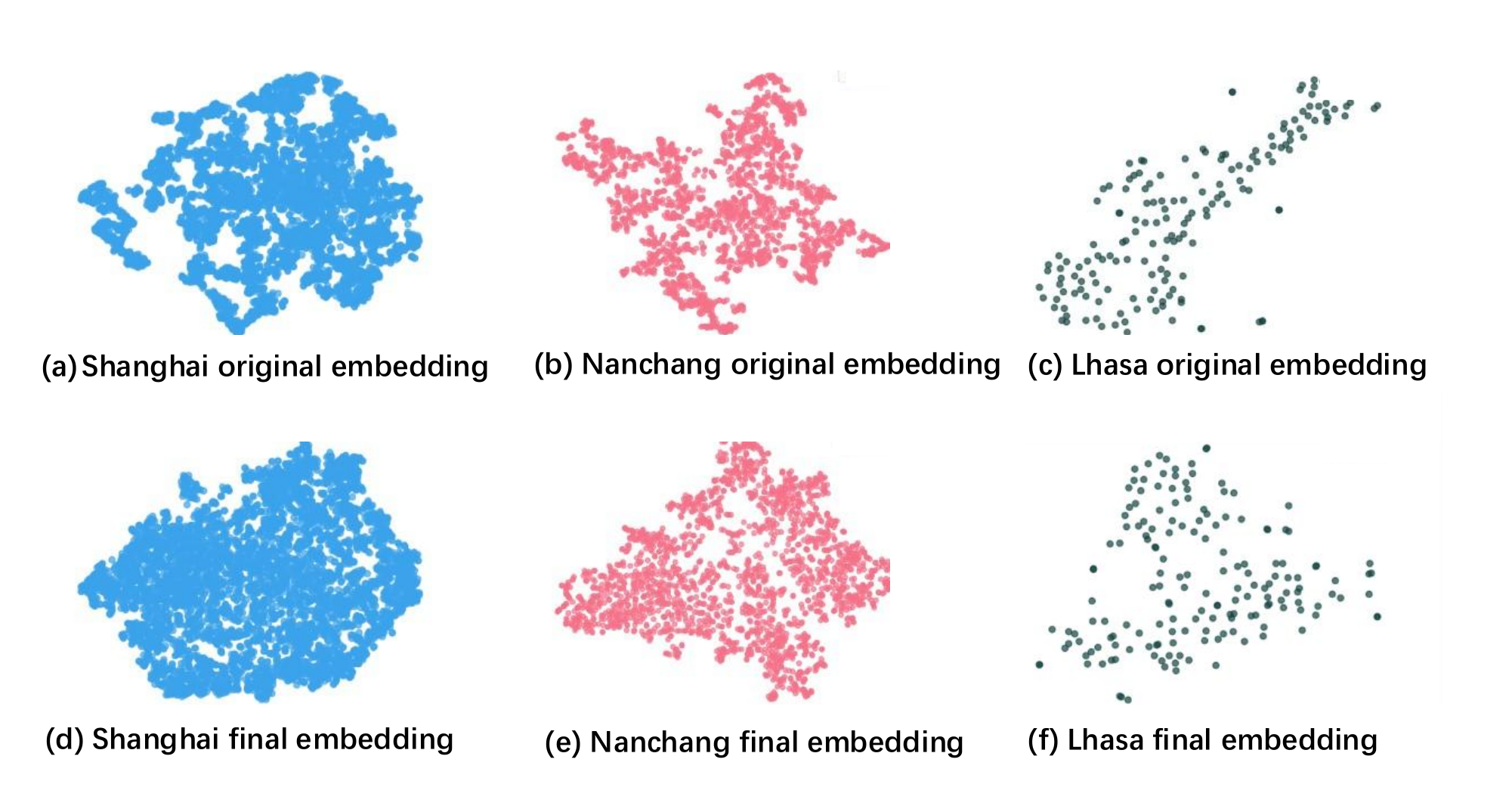}
    \caption{Locations embedding of three cities based on t-SNE visualization. (a)-(c) are original embeddings, and (d)-(f) are final embeddings learned by DCN.}
    \label{fig:embedding}
\end{figure*}

\subsection{Ablation Study}

To investigate the critical roles of different design modules in Mob-Former, we conducted ablation studies on three datasets. Specifically, we removed the feature-crossing DCN and the MoE layer from the model architecture. Since completely removing the DCN would lead to the model's inability to capture higher-order location features, we only removed the cross-layer of the DCN and replaced it with an MLP. For the MoE layer, we replaced it with an FFN (Feed-Forward Network) layer.

As shown in Table~\ref{tbl:ablation}, the experimental results indicate that when replacing the DCN with an MLP, the model performance deteriorated on datasets with a larger number of locations. This phenomenon demonstrates that the DCN has a unique advantage in capturing the differences between locations, allowing it to more accurately characterize the complex relationships between different locations, which is crucial for model performance.

Moreover, after removing the MoE layer, the performance degradation is more significant on the Nanchang dataset with a smaller data volume. This clearly indicates that the MoE layer plays a key role in handling the heterogeneity of data from different cities. It can effectively integrate data with similar mobility pattern trajectories, thereby enhancing the model's adaptability and generalization ability for cross-city data.

\subsection{Visualization of Location Embeddings}

The location tower extracts the representations of all locations within a city. Given that the distribution of location features varies significantly across different cities, to analyze how the location tower handles this heterogeneity, we select the original embeddings of each city after passing through the Location Encoder and the final location embeddings learned by DCN and use t-SNE dimensionality reduction for visualization.

Figure~\ref{fig:embedding} shows t-SNE visualization analysis of location embeddings for cities, it was observed that after learned by Deep \& Cross Network (DCN), the distribution of locations across different cities exhibited higher similarity. This indicates that DCN is capable of effectively extracting similar location feature patterns across different cities, thereby providing strong support for cross-city location analysis and modeling. Meanwhile, within the same city, the distribution of locations is no longer as concentrated as in the original embeddings, and the differences between various locations become more pronounced.

\begin{figure}[t!]
    \centering
   \includegraphics[width=\linewidth]{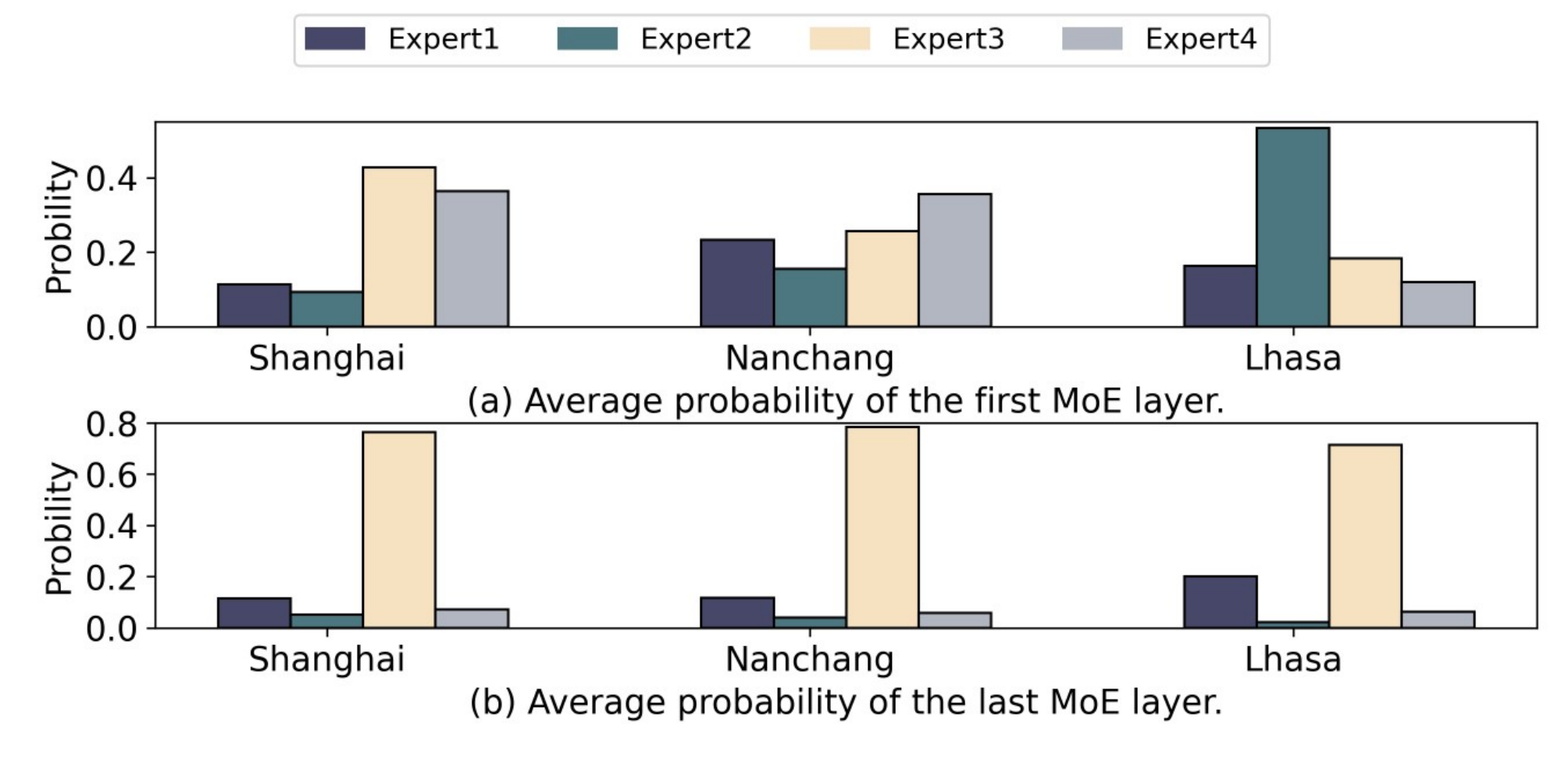}
    \caption{Expert selection distribution in MOE Layers across three city datasets.}
    \label{fig:moe}
\end{figure}

\subsection{In-depth Analysis of MoE}

To thoroughly investigate whether MoE can capture the common mobility patterns across different cities and to understand its mechanism for handling data heterogeneity, we have conducted a detailed statistical analysis of the expert selection probabilities of different MoE layers across various mobility data.

Figure~\ref{fig:moe} illustrate the average probabilities of the gating network outputs in the first and last layer for three datasets. It can be observed that there are significant differences in expert selection among various cities for the first layer, suggesting that the MOE dynamically selects the expert networks to be activated based on the different characteristics across cities. Meanwhile, the last layer exhibit more similar probability distributions, indicating that despite significant heterogeneity among different cities, similar patterns exist at higher-order dimensions. 

\begin{figure}[t!]
    \centering
    \includegraphics[width=\linewidth]{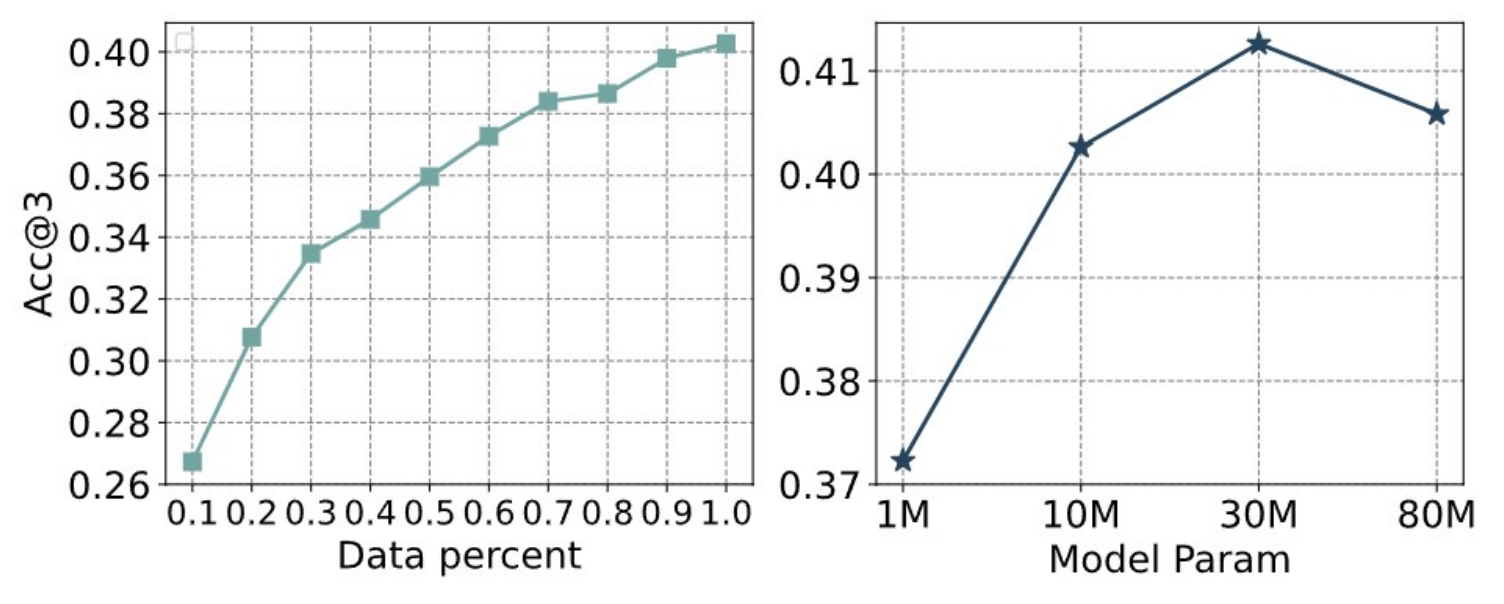}
    \caption{Data scalability(left) and model scalability(right).}
    \label{fig:scale}
\end{figure}

\subsection{Scalability}\label{scalee}

The high performance of UniMove is attributed to the joint training on multiple city datasets, which provides a richer information base. To verify the performance bottlenecks of the model under different parameter sizes and data volumes, we conducted experiments by testing the model's performance with varying parameter sizes and training data volumes.

Figure~\ref{fig:scale} shows the impact of varying model parameter scales and dataset sizes on prediction performance. We have identified two characteristics: (i) As the training data volume increases progressively, the model's prediction performance exhibits a continuous upward trend. However, once the data volume reaches a certain threshold, the performance improvement gradually slows down and stabilizes. (ii) When the model's parameter scale is increased, the performance improves incrementally. However, for the dataset used in this study, overly complex model structures lead to faster convergence, which in turn prevents the model from achieving optimal performance.

\begin{table}[t!]
\caption{Acc@1 results of the hyperparameter study.}
\label{tbl:anysys}

\begin{threeparttable}
\resizebox{0.8\columnwidth}{!}{
\begin{tabular}{cccc}
\toprule
Expert/Top k & Nanchang & Lhasa &Shanghai\\
\hline
4/2 &	0.338	&0.425&0.263\\
4/1 &	0.293	&0.403&0.255\\
4/3	&0.334	&0.426&0.263\\
3/2	&0.311	&0.412&0.260\\
6/2	&0.332	&0.426&0.273\\
\bottomrule
\end{tabular}}
\end{threeparttable}
\end{table}

\subsection{Hyperparameter of MoE Analysis}
In our exploration of how different hyperparameters within the Mixture of Experts (MoE) framework impact model performance, we conducted experiments by varying the number of experts as well as the top-k values. The experimental outcomes are displayed in Table \ref{tbl:anysys}.

Changing the number of experts alters the model's parameter. From the table, a reduction in the number of experts tends to have a detrimental effect on the model's performance. When the number of experts is decreased, the model's capacity to capture complex patterns and diverse features in the data is diminished, thereby leading to suboptimal results. Conversely, increasing the number of experts does not necessarily lead to further improvements in performance. This result is consistent with the analysis in Section \ref{scalee}. Due to the limited diversity and quantity of the current training data, adding more experts or model parameters does not enhance the model's performance. Furthermore, the influence of top k and the number of experts on model performance resembles each other. Under the current training data, reducing the top k value leads to a performance drop, while increasing it doesn't significantly boost model performance.

\begin{figure}[t!]
    \centering
    \includegraphics[width=1\linewidth]{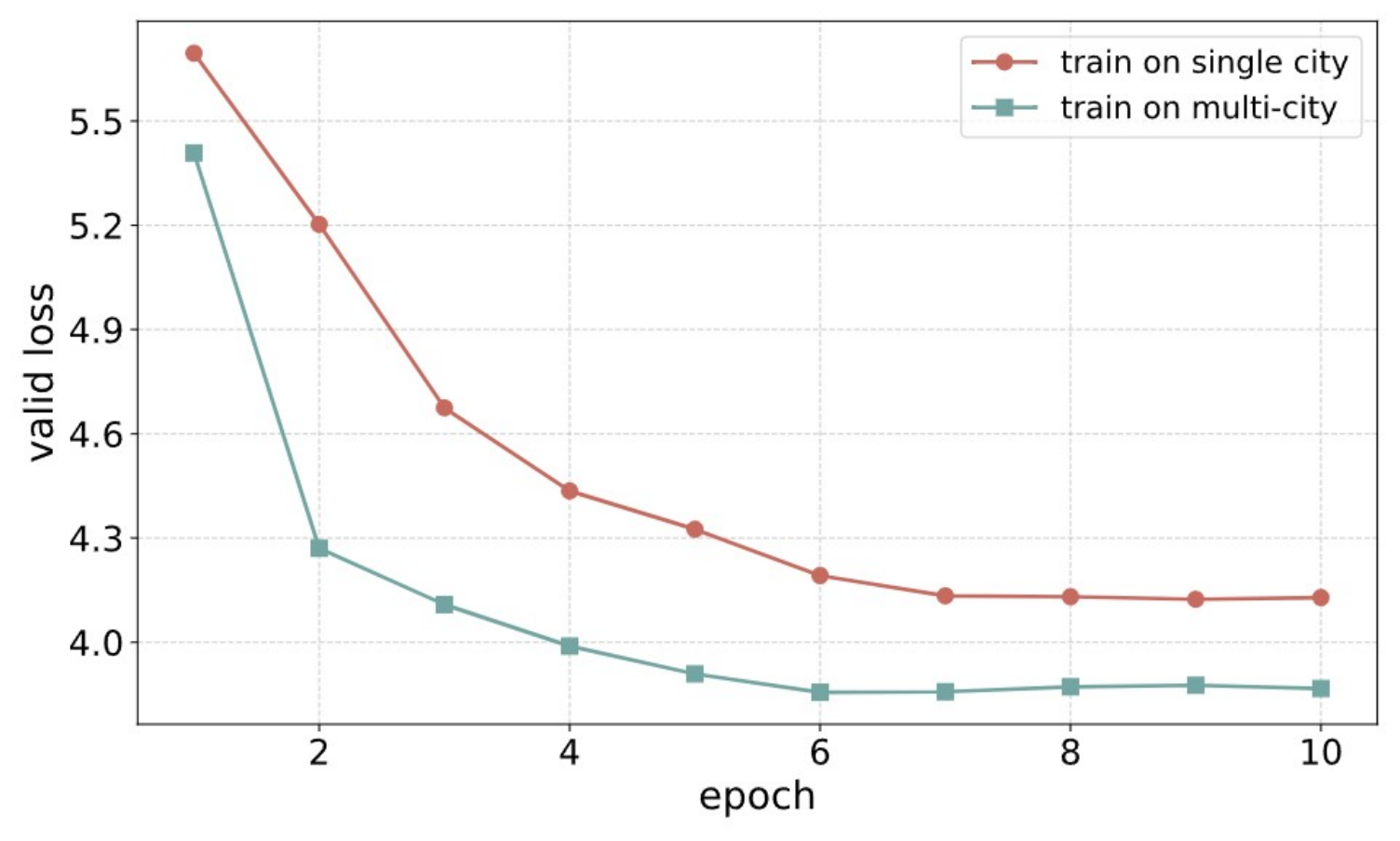}
    \caption{Comparison of training efficiency between single city and multi-city in terms of validation loss. }
    \label{fig:LOSS}
\end{figure}

\subsection{Training Efficiency}
To deeply explore if the multi-city joint training method can boost model performance and training efficiency, and whether it can help the model quickly grasp universal patterns while significantly improving prediction accuracy, we compared the validation loss of the UniMove trained on single city and multi-city datasets.
Comparing the validation loss of the UniMove model trained on single-city and multi-city datasets showed in Figure~\ref{fig:LOSS} , it demonstrates that joint training across multiple cities significantly enhances the training efficiency of the model. Specifically, within the same epoch, the reduction in loss value is much greater for multi-city training. The complementarity and similarity of data from different cities (such as variations in traffic patterns and user behaviors) provide a richer set of training information. This enables the model to rapidly capture universal patterns and achieve the optimal performance attainable by single-dataset training within fewer epochs. Moreover, it effectively improves the performance of prediction accuracy.

\section{Conclusion}

\vspace{+3mm}

In this work, we propose UniMove, a transformative approach to human mobility prediction that unlocks the potential for more robust and scalable modeling paradigms. By enabling joint training across multi-city data and promoting mutual enhancement across diverse urban environments, UniMove not only improves prediction accuracy but also advances the concept of a unified foundational model. Furthermore, the ability to leverage shared mobility insights across cities with varying structures and characteristics paves the way for more generalized and adaptable models that can be applied on a global scale.

While UniMove currently focuses on mobility prediction, its next-token prediction training framework lays a promising foundation for future expansion into other tasks. We plan to extend the model’s capabilities to include mobility generation, trajectory recovery, and anomaly detection. Additionally, this architecture holds the potential for integration with large language models, enabling a more seamless understanding of mobility data in natural language contexts. This synergy could facilitate more intuitive human-computer interactions and open new possibilities for urban decision-making.

\section*{Acknowledgments}
This work was supported in part by the National Natural Science Foundation of China under 62476152 and U24B20180.

\bibliographystyle{ACM-Reference-Format}
\bibliography{8.reference}





\end{document}